# Technological taxonomies for hypernym and hyponym retrieval in patent texts

You Zuo*, Yixuan Li**, Alma Parias García***, Kim Gerdes****

*INRIA de Paris, Paris, France
you.zuo@inria.fr
**Sorbonne Nouvelle University, France
yixuan.li@sorbonne-nouvelle.com
***Galytix, Prague, Czech Republic
almapargar@gmail.com
****LISN, CNRS and University Paris-Saclay, France
gerdes@lisn.fr

**Abstract.** This paper presents an automatic approach to creating taxonomies of technical terms based on the Cooperative Patent Classification (CPC). The resulting taxonomy contains about 170k nodes in 9 separate technological branches and is freely available. We also show that a Text-to-Text Transfer Transformer (T5) model can be fine-tuned to generate hypernyms and hyponyms with relatively high precision, confirming the manually assessed quality of the resource. The T5 model opens the taxonomy to any new technological terms for which a hypernym can be generated, thus making the resource updateable with new terms, an essential feature for the constantly evolving field of technological terminology.

Technological taxonomies for hypernym and hyponym retrieval in patent texts

# 1. Introduction

A patent application is a legal asset in text form that grants its owner the exclusive right to use the patented invention for a limited time. Companies and individual inventors are encouraged to fully disclose the technical knowledge embodied in their patented inventions to receive the benefits of greater intellectual property rights. Thus, patent publications are a good reflection of technological innovation and development worldwide.

Typically, patent applications are drafted by patent attorneys with technical and legal backgrounds on behalf of inventors. Patents are granted only if the claims present subject matter that is new and inventive relative to the prior art. Hence, already in this early stage of drafting a patent application, it is of primordial importance that the words and terminology chosen in the drafts are as general as possible to cover a broader scope while also mentioning specific cases to match more real-life application scenarios. It is a "play on words" with enormous economic importance. From this perspective, patent attorneys need to have an accurate understanding of the technical domain to cover the broadest possible semantic field surrounding the invention. Nevertheless, the patent domain is, by definition, at the forefront of technology, and most terms cannot be found in existing terminology databases. As a result, there is a tremendous need to meet the demand for taxonomies that include the most up-to-date technological expressions and that can be easily and continuously updated. Therefore, we decided to create a CPC-based taxonomy, specifically designed for the task of hyponym/hypernym retrieval of patent texts, which shares a large number of words with real patents and can be automatically updated every year.

CPC (Cooperative Patent Classification)[1] is an official patent classification system for technical documents, developed jointly by the world's largest patent offices: the EPO (European Patent Office) and the USPTO (United States Patent and Trademark Office), and today adopted and constantly updated by a broader consortium of patent offices. The CPC system is rich not only in the scale of terminological expressions at the frontiers of innovation but also in the relationships between technological expressions in the context of knowledge domains, with its hierarchic format. It is a taxonomy with a tree-like structure with five levels (as shown in the example in FIG. 1). It is firstly divided into the following nine sections A-H and Y, covering the vast majority of technological fields, plus a Y section to categorize new inventions for which there are not relevant categories yet. Emerging fields are inserted into the A-H taxonomy when a field stabilizes:

    A. *Human necessities*
    B. *Performing operations; transporting*
    C. *Chemistry; metallurgy*
    D. *Textiles; paper*

---

[1] https://www.cooperativepatentclassification.org/

E.  *Fixed constructions*
F.  *Mechanical engineering; lighting; heating; weapons; blasting engines or pumps*
G.  *Physics*
H.  *Electricity*
Y.  *General tagging of new technological developments; general tagging of cross-sectional technologies spanning over several sections of the IPC; technical subjects covered by former USPC cross-reference art collections [XRACs] and digests*

These nine sections are in turn subdivided at four levels: classes, sub-classes, groups, and sub-groups. Each node[2] in CPC has one or more headings in the form of noun phrases, participle phrases, or prepositional phrases, and there are over 250,000 nodes at the sub-group level of CPC.

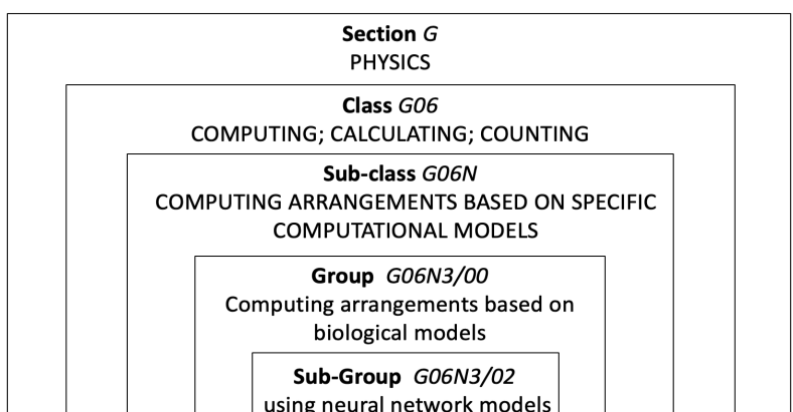

FIG. 1 – *Example of one CPC node G06N3/02.*

In this paper, we propose a heuristic rule-based approach for building domain-specific taxonomies of technical terms based on the CPC (Cooperative Patent Classification). We then test and evaluate different deep-learning models on the created taxonomies to obtain a model with enhanced knowledge of tech-taxonomy for predicting hypernyms/hyponyms for emerging terms or concepts (terms or concepts not in the created taxonomy but appearing in patent texts). The final system is a combination of the taxonomy stored in a database with neural network machine prediction. Experimentally, the system has proved useful for hypernym/hyponym retrieval, as well as for other text mining and information retrieval tasks for patent or scientific texts. Our created taxonomies and scripts are available for research purposes

---

[2] To distinguish it from the Class level in CPC, we refer to each unit in CPC as a node.

under the Creative Commons license CC BY-NC-SA 3.0 as detailed in the code repository.[3]

## 2. Related Work

### 2.1 Creation of Taxonomy or Ontology

Most of the existing taxonomies, ontologies, or semantic networks were designed for defining word meanings or structuring general knowledge, such as WordNet (Miller, 1995), FreeBase (Bollacker et al., 2008), BabelNet (Navigli and Ponzetto, 2012), and Wikidata (Vrandečić and Krötzsch, 2014) among others. Since they were not designed for domain-specific use, they contain a large number of expressions that are not relevant to science and technology and are difficult to filter out. Several works in medical and chemical taxonomy or ontology have been proposed, such as the biomedical ontology MeSH (Medical Subject Headings) whose English and French versions have been created for a thesaurus of vocabulary used to index articles for PubMed, and the Bio-chemical database and ontology of molecular entities focused on "small" chemical ChEBI (Degtyarenko et al., 2008). Outside the biomedical and chemical domain, technological taxonomies or ontologies such as NASA Technology Taxonomy [4] and (Oztemel et Gursev, 2020) on Industry 4.0 have also been proposed for specific engineering use. Another important resource is the Computer Science Ontology (CSO) (https://cso.kmi.open.ac.uk/downloads) that stores information about Computer Science research topics, the most up-to-date and exhaustive Computer Science topic ontology by far, that has been subject to semi-automatic extension attempts (Santosa et al. 2021). While these resources are of high quality in the relevant areas, they require a lot of manual work to maintain and update.

Some patent-related taxonomies or ontologies were created to help patent practitioners meet multiple needs. Patent ontology-related applications such as (Ghoula et al. 2007) (Wang et al. 2013) aim to build a patent semantic annotation system for patent document retrieval. (Pesenhofer et al. 2008) built manually a science taxonomy derived from the Wikipedia Science Portal[5] to associate relevant patents with particular Wikipedia pages. Later on, (Siddharth et al. 2011) (Siddharth et al, 2012) (Taduri et al. 2011) created patent ontologies specifically for structuring patent information from multi-resources such as patent documents, court cases, and file wrappers. (Inaba and Squicciarini, 2017) created the "J Tag" taxonomy (definitions of information and communication technologies) based on the International Patent Classification (IPC) technology classes to better align the definitions of their ICT (Information and Communication Technologies) sectors and ICT products into the patent terminologies. Using machine learning techniques, (Billington et al., 2020) construct a transparent, replicable, and adaptable patent taxonomy and a new automated methodology for classifying patents. Besides, (Sarica et al., 2020) created another knowledge semantic network "TechNet" from USPTO patent texts, which contains over four million engineering terms to inspire innovative design. However, none of them were interested in the hypernym or hyponym relations between technological expressions.

---

[3] https://github.com/ZoeYou/AutoTaxo
[4] https://techport.nasa.gov/view/taxonomy
[5] https://en.wikipedia.org/wiki/Category:Science_portals

We therefore decided to create a patent-related technological taxonomy based on CPC (Cooperative Patent Classification), which includes a large size of terminological terms and concepts in patent domains as well as rich hierarchical information between them.

### 2.2 Hypernym/Hyponym Prediction

Hypernym prediction[6] is a sub-task of relation prediction where the hypernymy denotes the IS-A relation that is used to create taxonomies of terms. The common test setup is to hide one entity from the relation triplet, asking the system to recover it based on the other entity and the relation type (IS-A in our case). Training a model to gain knowledge of created taxonomies is crucial for patent drafting in practice, as patent texts often contain new terms that may not appear in the taxonomy; therefore, we expect the model to be able to make inferences about new terms.

Early approaches such as (Weeds et al., 2014) and (Vyas and Carpuat, 2017) consider the task of hypernym prediction as a binary classification (hypernym detection) of whether two given words or multi-word expressions are in a hypernym relation. Later solutions such as (Yamane et al., 2016), (Ustalov et al., 2017), and (Bernier-Colborne and Barrière, 2018) proposed supervised projection learning methods to learn multiple matrices that project a query embedding such that the projection is close to its target hypernym. Other approaches to hypernym prediction were primitively designed for knowledge base completion, where hypernym is considered as one of the semantic relations between two nodes in a graph. The pioneering work in this area is TransE (Bordes et al., 2013), various approaches have been proposed later to improve different parts of the learning architecture as DistMult (Yang et al., 2014), TransH (Wang et al., 2014), TransR (Lin et al., 2015), TransD (Ji et al., 2015), etc. Later methods were proposed using more sophisticated deep learning networks or modeling strategies. ConvKB (Nguyen et al., 2017) proposed a novel embedding model that applies the convolutional neural network to explore the global relationships among same dimensional entries of the entity and relation embeddings. M3GM (Pinter and Eisenstein, 2018) created a method which extended the Exponential Random Graph Model (ERGM) that scales to large multi-relational graphs; by combining global and local properties of semantic graphs, it substantially improves performance on link prediction. (Cho et al., 2020) formulated the hypernym prediction as a sequence generation task, they trained an LSTM-based model to predict the hypernym of the given input or the previous prediction in the output sequence.

---

[6] We do not discuss related research about hyponym prediction because 1) it is a symmetric task for hypernym prediction; 2) hypernym prediction is easier to be formulated mathematically since each entity should have only one hypernym in a well-defined taxonomy.

The emergence and increasing use of transfer learning methods in natural language processing in the past few years have also shown their effectiveness in various methods, methodologies, and practices. The textual encoding method KG-BERT (Yao et al., 2019) fine-tuned a pre-trained encoder BERT (Devlin et al., 2018) to concatenate triples' text for deep contextualized representations. StAR (Wang et al., 2021) applied a Siamese-style textual encoder to the triple for two contextualized representations, with two parallel scoring strategies used to learn both contextualized and structured knowledge. However, pre-trained generation models have yet to be explored in the hypernym prediction task. In our study, we explore the seq2seq pre-trained language generation model T5 (Text-to-Text Transfer Transformer) (Raffel et al., 2020) to test its transfer learning capabilities in hypernym discovery.

## 3. Technological Taxonomy Creation

### 3.1 Original Form of CPC Titles

The original data we use come from the latest version of CPC titles.[7] For each field at the section level (A-H, Y), it provides a separated text file with three columns: the CPC codes, the ranking of sub-groups, and the CPC titles. The second column exists in the sense that although the official definition of CPC has only five levels (section, class, sub-class, group, and sub-group), in practice there are often more subdivided hierarchical relationships within the most granular level, the subgroup, with the deepest subgroup reaching up to 12 levels. As in the example in Table 1, the numbers in the second column indicate the level of the subgroup level, with 0 indicating that the title belongs to the group level, and 1, 2, and 3 indicating respectively that the title belongs to the first level, the second level, and the third level of the subgroup, etc.

| G | | PHYSICS |
|---|---|---|
| G06 | | COMPUTING; CALCULATING; COUNTING |
| G06N | | COMPUTING ARRANGEMENTS BASED ON SPECIFIC COMPUTATIONAL MODELS |
| G06N3/00 | 0 | Computing arrangements based on biological models |
| G06N3/02 | 1 | using neural network models |
| G06N3/06 | 2 | Physical realisation, i.e. hardware implementation of neural networks, neurons or parts of neurons |
| G06N3/063 | 3 | using electronic means |

---

[7] https://www.cooperativepatentclassification.org/sites/default/files/cpc/bulk/ CPCTitleList202202.zip

| G06N3/0635 | 4 | *{using analogue means}* |

TAB. 1 – *Examples of original CPC titles from text file of class G.*

We use the CPC class titles as a data source for our taxonomy building because they contain a large amount of terms as well as structural information (relations between units). More than 60% of them contain lists, coordination, or disjunction, and more than 20% of them contain one or more terms followed by expressions like "e.g."/ "such as" to indicate special cases or what comes after "i.e." or with square brackets to indicates synonyms. Several of the biggest challenges of building tech-term taxonomy according to CPC stem from the facts that 1) CPC titles are not full category names, but are supplements to their parent category titles, adding new information (e.g., in Table 1., the title of G06N3/063: "using electronic means" is a supplement to its parent category G06N3/06); 2) pronominal references to its parent category or previous content in the same title marked with "thereof", "therefor", "therewith", etc.; 3) some CPC titles are not descriptive but refer exclusively to their adjacent categories, such as CPC category G01M99/00 with its title "Subject matter not provided for in other groups of this subclass".

In the next sections, we propose the title2term algorithm to address these challenges. It is worth noting that some entries in our taxonomy are terms in the sense of "specialized linguistic units that represent domain concepts" (Roche et al., 2009; Suonuuti, 1998), while others are descriptive intermediate elements, but are still correct entries in our taxonomy. Each unit in our taxonomy is a designation that represents a general concept by linguistic means.[8]

## 3.2 Algorithm of title2term

We build a rule-based algorithm for converting English CPC titles into nine domain-specific taxonomies. Each CPC text file is first converted and saved in a strict tree structure that sorts its title nodes according to the hierarchy of the original CPC. We maintain the tree structure of the CPC system throughout the pre-processing of the data and the construction of the taxonomy.

The principal rules that we implemented in our work can be summarized in the following steps:

I. Text pre-processing
   In this step, we clean the irrelevant information and remove useless nodes for technical taxonomy.
   a. Delete contents containing CPC codes with round brackets, for example:

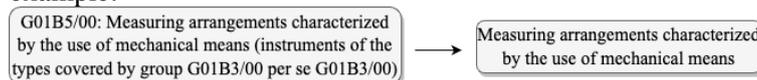

---
[8] [ISO 1087-1] : https://www.iso.org/obp/ui/#iso:std:iso:1087:ed-2:v1:en:term:3.4.1

- b. Remove braces
  CPC content with braces indicates that the content does not appear in the IPC, but the flower brackets do not give us any information for the taxonomy build.

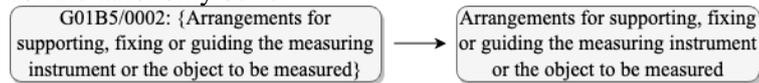

- c. Check if there are CPC codes within the node, and if the condition is met, delete the title and all its sub-titles. For example:

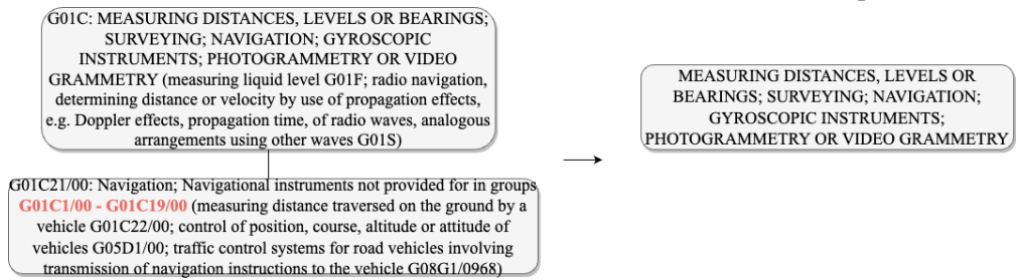

II. Node splitting
   After cleaning up all the useless information, we split the CPC titles into units in the taxonomy.
   - a. Split by semicolon
     Split parts will become sibling nodes and be connected to the same parent node, and inherit the same sub-nodes.

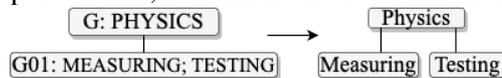

   - b. Split by "e.g."/ "such as"
     The content after "e.g."/ "such as" refers to an example of the content before it, in which case we consider this example to be a hyponym.

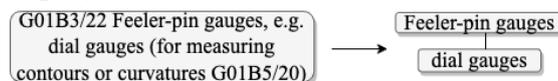

III. Replacements and attachments
   - a. Replacement of "such"
     Here "such" refers to one of the previously mentioned things, so when splitting the title, we simultaneously replace "such" with the object it refers to, for example:

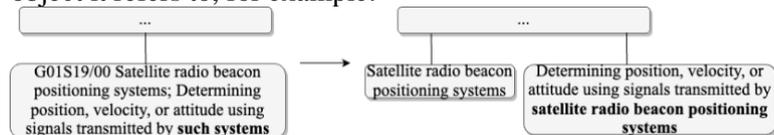

   - b. Replacements "thereof", "therewith", "therefor"

For CPC titles that end with these adverbs, they usually modify the previous content (possibly in the same title's previous part, or in their parent title). In this case, we replace firstly "thereof", "therewith," and "therefor" with "of", "with," and "for", then append to them what they modify. For example:

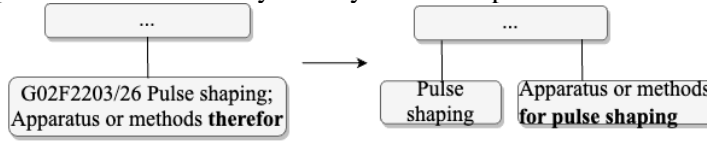

c. CPC titles starting with a lowercase
In CPC entry files, a title that begins with a lowercase letter means that it complements its parent title and therefore needs to be linked to it.

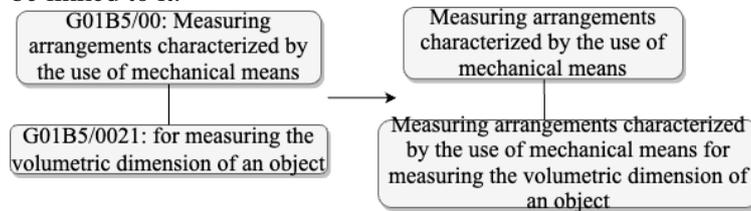

d. CPC titles starting with "details" or "Subject matter not provided for in other groups", etc.
CPC categories with the title of "details", and "details of xxx" do not themselves provide information in taxonomy, but they imply that its subcategories can be identified as belonging to its parent category. In this case, we delete this category node and connect all its subcategories to its parent category node. For example:

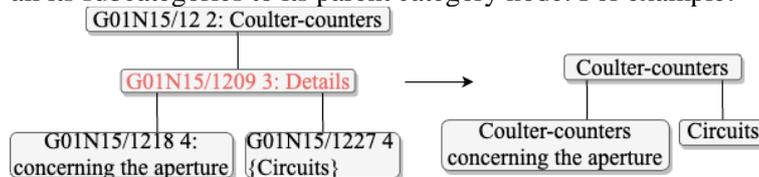

IV. Synonym Extraction
In our taxonomy, we do not present synonym relationships, but we still extract them and save them in an additional file for later use. The synonyms are indicated by abbreviations in square brackets, or the content follows the "i.e." in the CPC titles.

  a. Content in square brackets
  For example, for CPC category G01N2021/5903 with the title *"{using surface plasmon resonance [SPR], ..."* "SPR" is synonymous with *"surface plasmon resonance"*.
  b. Content following "i.e."

> For example, for CPC category G01C21/3438 *"{Rendez-vous, i.e. searching a destination where several users can meet, and the routes to this destination for these users; ..."*, *"searching a destination where several users can meet, and the routes to this destination for these users"* is saved as synonyms of *"Rendez-vous"*.

### 3.3 Statistics of Taxonomies

After applying the data pre-processing and title2term algorithm on titles of each domain section, the total number of term-hypernym pairs we have is as in Table 2. Note that is very uneven across sections since we restricted CPC titles only to the second level of sub-group and the reduction and the CPC sections differ in the number of sub-categories at deeper levels.

| Domain (CPC sections) | # Titles in original files | # Term–hypernym pairs in taxonomy |
|---|---:|---:|
| A. Human necessities | 29 650 | 25 551 |
| B. Performing operations; transporting | 56 503 | 40 033 |
| C. Chemistry; metallurgy | 38 243 | 33 232 |
| D. Textiles; paper | 5 691 | 4 005 |
| E. Fixes constructions | 9 248 | 7 517 |
| F. Mechanical engineering; lighting; heating; weapons; blasting engines or pumps | 27 979 | 17 962 |
| G. Physics | 37 839 | 17 697 |
| H. Electricity | 39 137 | 15 237 |
| Y. new technological developments | 16 186 | 8 852 |
| **TOTAL** | 260 476 | 170 086 |

TAB. 2 – *Number of class titles in CPC original files and number of term-hypernym pairs in created taxonomies (since we limited CPC titles to the second sub-group level of and also because of step I.c above, numbers in the third column are always inferior to the second column).*

## 4. Evaluation

### 4.1 Manual Evaluation

To review the quality of our taxonomy, we manually evaluated a list of 200 term-hypernym pairs randomly selected from the created taxonomy. Limited in time and knowledge, we chose pairs only from section G that the authors are most familiar with. We considered two aspects of its precision 1) the expression itself and 2) the triplet (relation of the two terms). Among the 200 term-hypernym candidates, we noticed 11 problematic expressions, 4 of which are incorrect for long expressions because of the attachment step III, as an example of the detected problems consider *"methods or arrangements for sensing record carriers by electromagnetic radiation **sensing** by electromagnetic radiation **sensing** by radiation using wavelengths larger than 0.1 mm arrangements for protecting the arrangement comprising a circuit inside of the interrogation device"*. We see that the multiple attachment steps created a highly complex expression that is not a term in a classical sense. Other errors stem from ambiguous expressions such as *"Coherent methods,"* or *"Heads"* which are too general and are not actual hyponyms. Concerning semantic relations, 170 of the 200 pairs can be qualified as term-hypernym pairs, which is a precision of 85%. Among the errors, more than half are pairs of an instance and a process or action (e.g., *"Testing, calibrating, or compensating of compasses"* whose hypernym is not really the proposed *"Compasses"*); other cases involve problematic expressions and inversions of the relation (term pairs should be in a hypernym relation but is in a hyponym relation) such as *"Saddling equipment for riding or pack-animals"* being the hyponym of *"STIRRUPS"*.

### 4.2 Automatic Evaluation

We fine-tuned the t5-base[9] seq2seq model in its PyTorch version from huggingface for hypernyms /hyponyms prediction with the following hyperparameter settings: optimizer = AdamW, learning rate = 1e-4, max length = 128, batch size = 16, and number of epochs = 6. The input format of the T5 model is fixed as "predict hypernym: " / "predict hyponym: " + <special token of domain> + term expression. For the output generation, we also set max length = 128, and beam number and beam size are both set to 10. We trained two models, one for predicting hypernyms and the other for predicting hyponyms for a given expression in its corresponding section domain, respectively. For the dataset, we extracted and mixed all term-hypernym pairs from each domain, and then split the training and test data at a ratio of 0.8 and 0.2. Two different metrics are applied to evaluate the accuracy of model predictions: hits@k (k=1, 3, 10) and MRR (mean reciprocal rank). Hits@K represents the ratio of test instances with correct candidate terms ranked top-k, and mean reciprocal rank (MRR) reflects the absolute ranking.

| Model | Hits@1 | Hits@3 | Hits@10 | MRR |
|---|---|---|---|---|
| term->hypernyms | .2986 | .3705 | .4410 | .3516 |
| term->hyponyms | .3014 | .3675 | .4402 | .3516 |

TAB. 3 – *Model performances of T5 fine-tuned on our term-hypernym / term-hyponym pairs.* The two models for predicting hypernyms and hyponyms obtained similar performance, and on average, there was a 40% chance that the model could predict the correct outcome in the top 10 prediction. We also did an ablation study to test the

---

[9] https://huggingface.co/t5-base

enhancement effect of special tokens on the model, the improvement in Hits@10 is 1.19%.

In the following table, we show the different predictions for the same input with different domain special tokens, demonstrating that the T5 model knows to distinguish the semantic meaning between different domains. We can see that for domain A the model is more biased toward predicting human-computer interaction; for domain C (Chemistry), where the term "audio feedback" is very rare, the model gives fewer convincing proposals in the domain of environmental measurement and control, while for domain G, the model gives good predictions of specific devices.

| Domain | Input | Predictions |
|---|---|---|
| A. Human necessities | Predict hypernym: <A> audio feedback | input arrangements for interaction between user and computer |
| | | input arrangements for interaction between player and computer |
| | | user input interfaces for electrophonic musical instruments |
| C. Chemistry; metallurgy | Predict hypernym: <C> audio feedback | characteristics or properties of obtained polyolefin |
| | | means for regulation, monitoring, measurement or control |
| | | feedback signal in controlled environment |
| G. Physics | Predict hypernym: <G> audio feedback | sound-producing device |
| | | feedback to the output device |
| | | feedback to the audio signal in a recording device |

TAB. 4 – *Examples of T5 model's ability for domain-specific hypernym prediction.*

## 5. Conclusion and Future Work

To conclude, in this paper we have shown how the well-curated patent classification system CPC can be used as a resource for developing 1) an open high-quality taxonomy of technical terms and 2) a T5-based hypernym generator that allows for validation of the coherence of the taxonomy as well as for hypernym/hyponym generation. We project the use of such a system in the patent draft process where it can propose more general (hypernyms) or more specific (hyponyms) terms for a given term, and where it can allow adding variants of the claims into the description, a common practice that allows inventors and their attorneys to extend the scope of the patent applications.

For future work, we plan to introduce syntactic features such as POS-tagging and dependency parsing to better split CPC titles because some of our taxonomy entries are still disjunctions, such as "Potatoes, yams, beet or wasabi" that should be separated and integrated as individual units. In addition, we will try to convert and preserve our taxonomies in specialized ontological software such as Protégé.[10] Note that the CPC is a moving target as it is constantly updated by the patent offices, with new classes reflecting the need to classify emerging technologies. The ontology that we developed and the corresponding code is made available for research purposes under the Creative Commons license CC BY-NC-SA 3.0 as on https://github.com/ZoeYou/AutoTaxo and will continuously be improved and updated.

---

[10] https://protege.stanford.edu/

Technological taxonomies for hypernym and hyponym retrieval in patent texts